# A comprehensive Persian offline handwritten database for investigating the effects of heritability and family relationships on handwriting

Abbas Zohrevand[a], Javad Sadri[b,c,*], Zahra Imani[a]

[a] Department of Computer Engineering, Kosar University of Bojnord, P.O. Box : 9415615458, Bojnord, Iran

[b] Department of Computer Science & Software Engineering, Faculty of Engineering and Computer Science, Concordia University, Montreal, Quebec, Canada, H3G 1M8

[c] Department of Computer Engineering, Faculty of Electrical and Computer Engineering, University of Birjand, P.O. Box: 615/97175, Birjand, Iran

## Abstract:

This paper introduces a comprehensive database for research and investigation on the effects of inheritance on handwriting. A database has been created that can be used to answer questions such as: Is there a genetic component to handwriting? Is handwriting inherited? Do family relationships affect handwriting? Varieties of samples of handwritten components such as: digits, letters, shapes and free paragraphs of 210 families including (grandparents, parents, uncles, aunts, siblings, cousins, nephews and nieces) have been collected using specially designed forms, and family relationships of all writers are captured. To the best of our knowledge, no such database is presently available. Based on comparisons and investigation of features of handwritings of family members, similarities among their features and writing styles are detected. Our database is freely available to the pattern recognition community and hope it will pave the way for investigations on the effects of inheritance and family relationships on handwritings.

# 1. Introduction

Inheritance influences many characteristics, behaviors, and functions of humans [1]. This is especially true for physical characteristics, including fingerprints [2], face shapes [3], iris [4], hand geometry [5,6], and voice [7]. Behaviors, such as walking, talking, drawing signatures [8], and handwritings [9], are widely used in applications related to individual identification, understanding of personality traits, and others. Among these features and behaviors, handwritings offer much information about the personality, identity, and health condition of their writers. Research suggests that certain physical aspects contributing to handwriting, such as hand-eye coordination, motor skills, and even hand bone structure, have genetic influences. These factors can create subtle family resemblances in handwriting, especially when considering baseline factors like letter size and slant. For example, muscle memory and fine motor control, both influenced by one's anatomy, play a role in how handwriting develops, thus giving a genetic component to some extent[10,11]. However, some handwriting characteristics are acquired through learning and practice. This includes not only the way letters are formed but also how consistently one writes. Factors such as the individual's educational background, the handwriting style they were taught, and their level of practice significantly shape how one's handwriting evolves over time[12]. For research purposes, handwritings are commonplace, low cost, accessible, and may be collected from subjects of any age or in most health conditions. This is in contrast to the challenges of gathering data on physical characteristics, such as iris, fingerprints, or voices, whose capturing required specialized and costly devices.

Still unanswered, however, is the question of whether inheritance and family relationships affect handwriting. In many families, handwritings of children, parents, and grandparents have similarities or share several common features in writing styles. For example, some family members may write particular words or letters with similar slants or the words and texts written by some may match those of others in size, shape, or texture. Fig.1 provides an example. Knowing the effects of inheritance on handwritings can improve many critical applications. These include applications that increase the performance and reliability of available systems for writer or signature identification or verification, handwritten recognition, adaptability of hand recognition systems to writers, and examining the authenticity of historical documents (for historical document examinations). For instance, the members of one family may exhibit the same slant feature in their handwritings. Thus, if a writer identification system uses this slant as its main feature for recognizing individual family members, the result will be a low performance that yields many false positive results. On the other hand, by examining the handwritings of a historical family, one can verify whether a certain document claimed to have been written by one of the family members is actually forgery or genuine. Knowing the effects of inheritance on handwritings can significantly or adaptively improve the feature selection process. Through this investigation, the accuracy, reliability, and robustness of all corresponding systems for writer identification systems can improve.

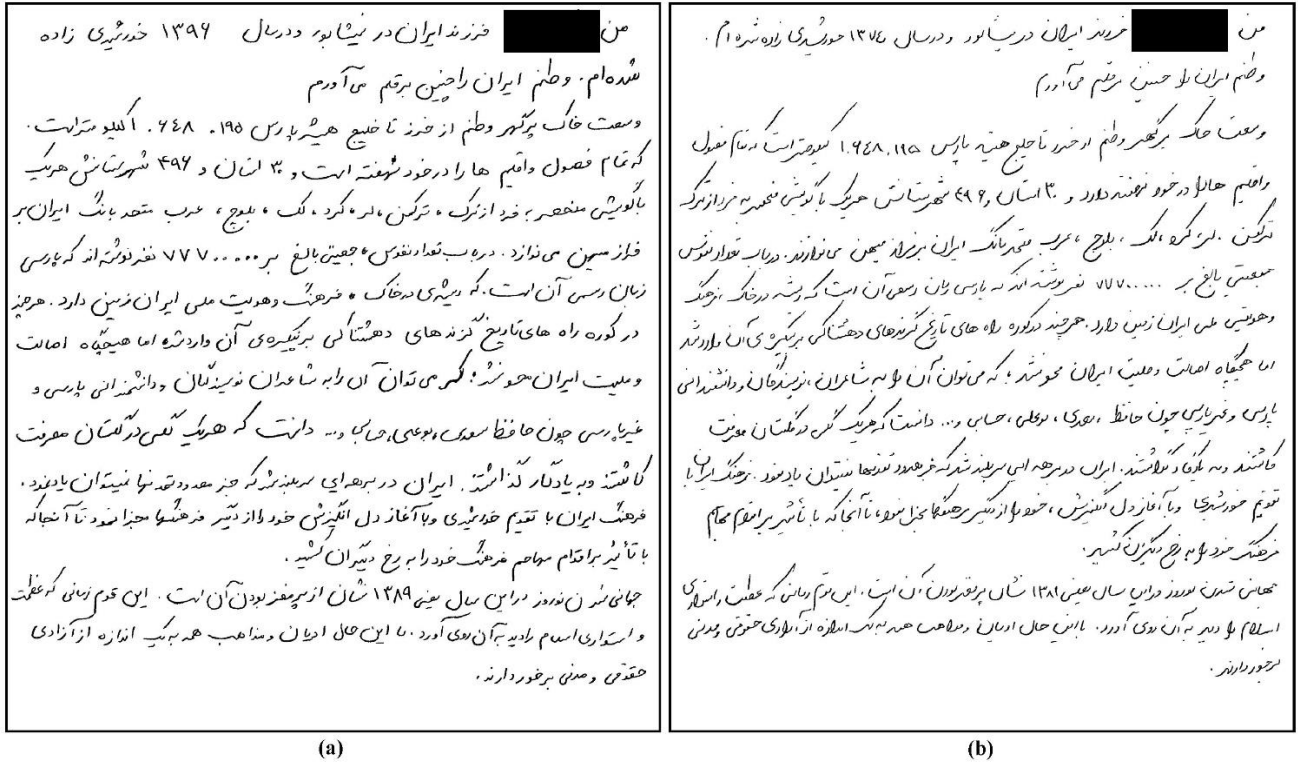

Fig. 1. (a), (b), Two samples of handwritings of the same text written by two cousins in Persian script. There are similarities in texture, word size, and slants.

To the best of the present study's knowledge, no comprehensive handwritten database has been created for investigating the effects of inheritance and family relationships on handwriting features of family members. Many databases have been developed for evaluating recognition systems and prediction models in pattern recognition or machine learning communities. For example, databases, such as for fingerprints in [13] and faces in [14], have been provided for identification or recognition purposes. For verification tasks, other databases have been created, such as for signatures in [15,16], which are utilized in plenty of applications for authenticating bank checks, contracts, and other security documents [17–19]. As mentioned, human handwritings have unique and varied properties, so there is a broad range of applications, such as gender identification [20,21], document question examination [22], individuality of handwriting [23][24], and writer identification and verification based on handwriting documents[25–27]. Therefore, depending on the script, several handwritings databases have been created, such as for Latin [28–30], Chinese [31], Korean [32], Japanese [33], Indian [34], Arabic [35,36] and Persian [37–40]. While these works are impressive endeavors to create databases, they are not useful for investigating the effect of inheritance on handwritings because none record the family relationships among the writers.

The main contributions of the present article are two-fold. Firstly, it creates a large-scale database containing different elements of handwritings such as: digits, alphabets, shapes, and free texts for investigation the effects of inheritance and family relationships on the handwritings of families. Secondly, for the purposes of providing handwriting databases to research the effects of inheritance and family relationships on handwritings, the current work introduces a generalized framework that can easily be followed by researchers working on handwritings of any script. This dataset was gathered from handwriting samples of all members and relatives of 210 families with an average size of 10 persons each, which were randomly selected from various geographical regions and varying educational and age levels living in Iran. This novel database not only provides excellent opportunities for conducting studies on classical problems on handwriting such as, shape recognition, skew correction, line segmentation/detection, word spotting/segmentation/recognition, writer identification, and gender, age and handedness detection, it also opens new windows for the researchers in pattern recognition community to explore the effects of inheritance and family relationships on handwriting at different levels.

The rest of the current paper is organized as follows. Section 2 describes the data collection process while Section 3 explains data extraction from Handwriting Sample form (HSFs). An overview of our database is presented in Section 4 and Section 5 explains some initial experimental results on this database. Section 6 compares the present work's database with that of similar works and finally Section 7 provides a conclusion and mentions future works.

## 2. Data collection process

Unlike conventional handwriting databases, the current study chose the family relationships among the writers as the key elements in the creation of this database. Thus, coding family relationships was a critical challenge in data collection and data organization. As seen in Fig. 2, each family is presented in a family tree, in which each family member is an individual node and the relationships between two nodes or family members are shown by edges. 210 volunteers were selected as representatives and they were trained for data collection. Each volunteer or so-called center was asked to collect handwritten sample forms of his/her immediate family members and relatives. Within each family, those members having no common genetic roots with the corresponding center were excluded from the data collection process. In Fig. 2, these excluded members have red dotted rectangles. Since family relation coding is one of the most important sub-tasks in our data collection, the next subsection explains the relationship coding method and the subsections afterwards describe the details of the data collection process.

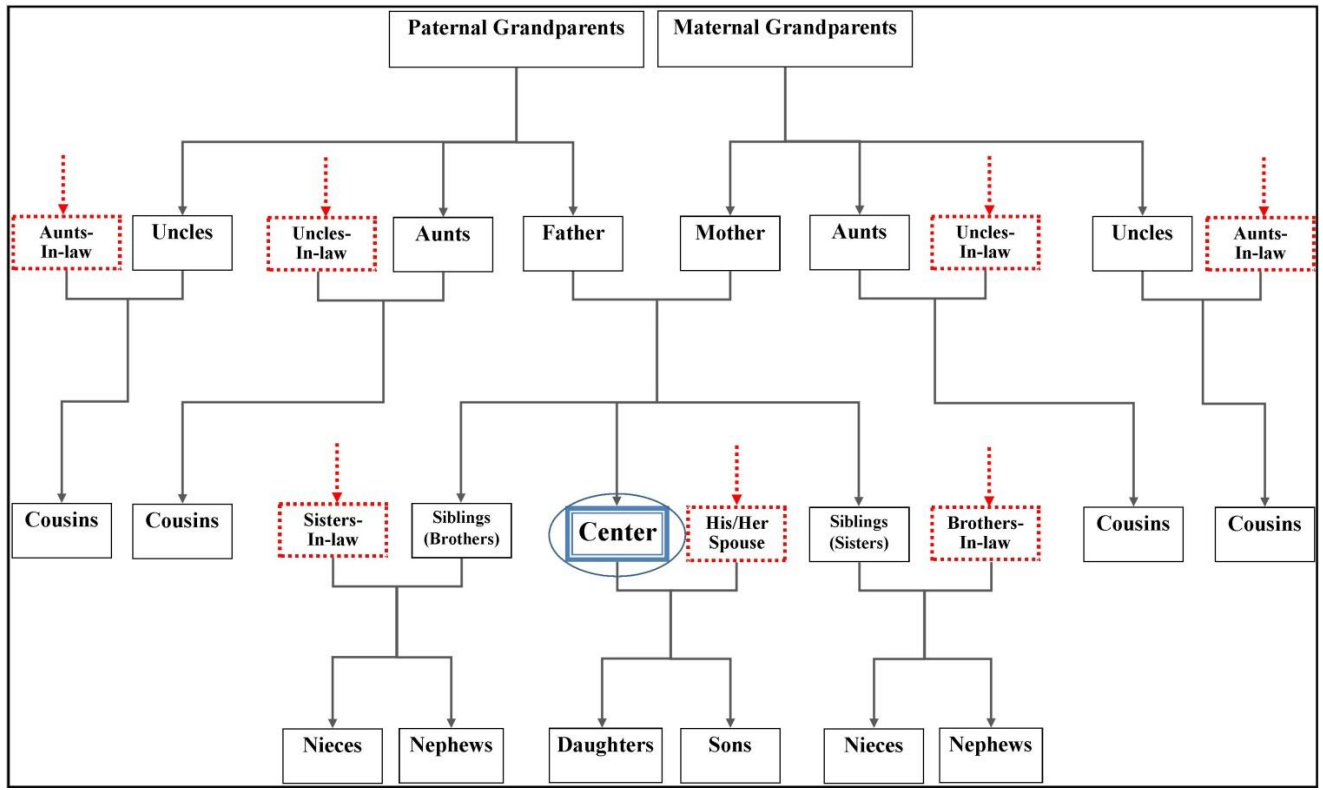

Fig. 2. The typical family tree considered in data collection. The center is in the rectangle with the blue border and all relationships are encoded with respect to the center. The red dotted rectangles show those members having no common genetic root with the corresponding center and these are excluded from the data collection.

## 2.1. Family relationship coding schema

The center in each family is the key-point for the family relationship coding inside the family. Therefore, if the tree shown in Fig. 2 is hung from the center node, then the layout of the tree changes and the center will become the central node in the revised tree. Then all other family members (other nodes) which have a relation with the center are arranged around the center (see Fig. 3). As shown, the tree in Fig. 3 has some important properties:

- If this tree is hung from the paternal and maternal grandparents, the original tree in Fig. 2 can easily be recreated.

- There are unique forward paths from the center to all other nodes (all his/her family members).

- These forward paths can be uniquely encoded into family relationship codes with respect to the center (the details of these codes are explained below).
- These family relationship codes allow all forward paths to be determined and so this tree can be uniquely reconstructed.

As illustrated in Fig. 3, each unique forward path from the center to all other family members can be simply encoded by some digits and two special characters (see Table 1). For example, the mother of the center is encoded by "0_1.1." In this code, "0" indicates the center, "1" shows mother, and ".1" denotes the multiplicity of the mother, which is one. For example, if the center has n brothers or n sisters, these different brothers or sisters are uniquely assigned codes such as: "0_4.1", "0_4.2", ... , "0_4.n" or "0_3.1", "0_3.2", ... , "0_3.n" respectively. The "_" shows the transition between family relationship levels and the number of "_" s determines the degree (distance) with respect to the center of each family. As seen in Table 1, only first-degree family relationships with respect to the center are encoded. The other levels and hierarchy of family relationships are encoded by combining the same codes for uncles, aunts, cousins and so on. For example, the first uncle of the center is coded as the brother of the father of the center: "0_2.1_4.1". Another example is the coding of second son of the third uncle of the center: "0_2.1_4.3_6.2.".

Table 1. Family relationship coding schema.

| Name | Code |
|---|---|
| Center (him/herself) | 0 |
| Mother | 1 |
| Father | 2 |
| Sister | 3 |
| Brother | 4 |
| Daughter | 5 |
| Son | 6 |
| Wife | 7 |
| Husband | 8 |
| Transition between levels | _ |
| Multiplicity indicator | .(dot) |

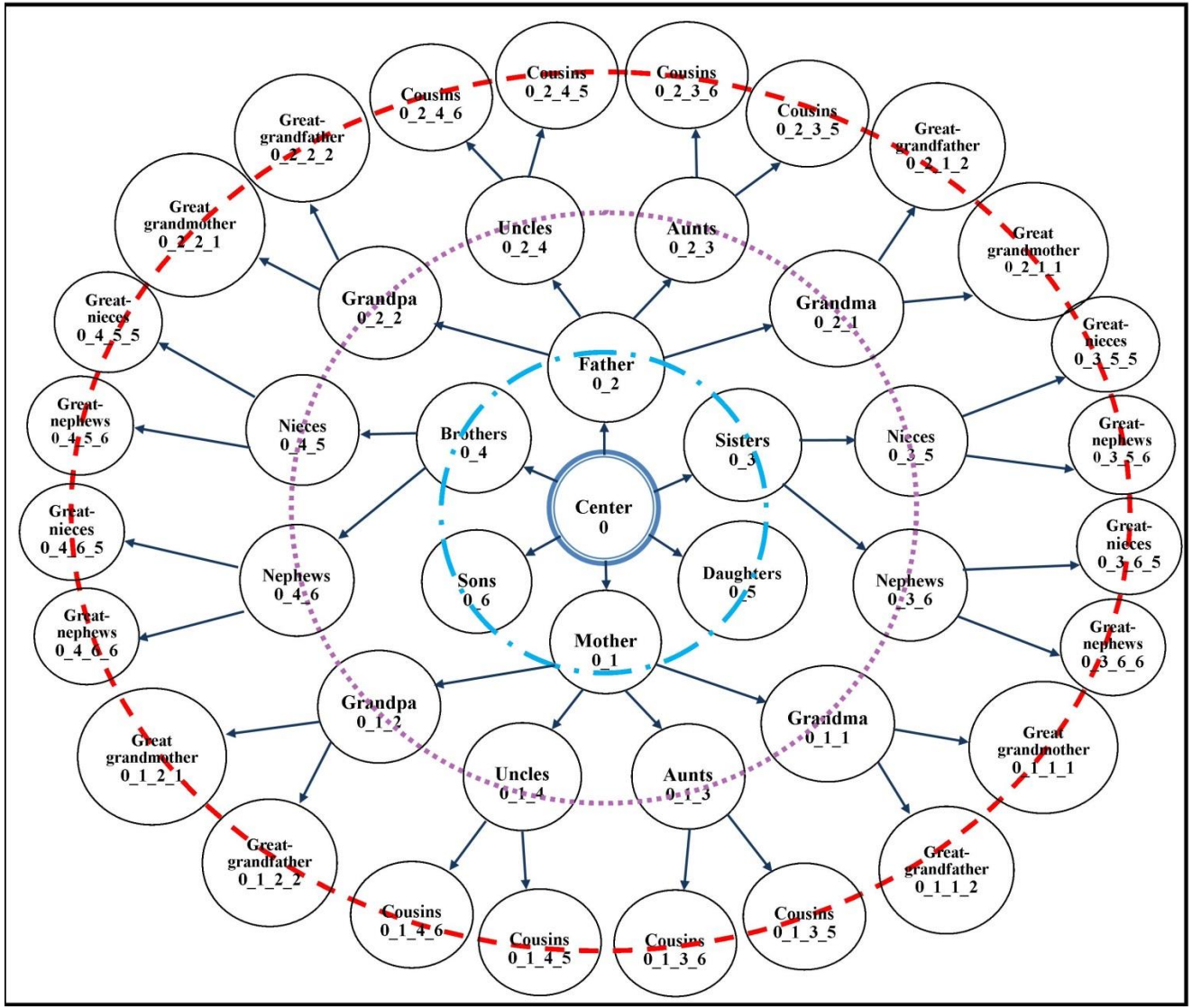

Fig.3. A tree with a central node (center of family), which presents the family relationship schema for three levels of family relations.

## 2.2. Design of Handwritten Sample Forms (HSFs)

As seen in Fig. 4, a two-page form was designed, each containing header and data blocks. A small black box was drawn on the corners of each form for the de-skewing process of the forms after scanning as well as for finding the location of the information for segmentation purposes. For de-skewing the forms' images after the scanning process, a small black box was drawn on the corners of both forms (F5 in Fig. 4a and Fig. 4b). For filling out the HSF forms, there no limitations were imposed on writing styles or writing instruments. The layout of each page has two blocks consisting of the header and an area below containing the data entry block(s). A completed sample of Page #1 and Page #2 are provided in Fig. 4a and Fig. 4b, respectively.

Fig. 4. The layout of Pages #1 and #2 (a and b) data collecting forms. Each page has two parts: the header (ground truth data) and data (samples of written items). According to the present study's confidentiality agreement, the writer's name was been blocked.

The header block includes the following information in the writer's Ground-Truth (GT) data: first name, last name, gender, handedness, age, and education level (F1 in Fig. 4a and Fig. 4b); in addition, the family relationship with respect to the center is captured (F6 in Fig. 4a and Fig. 4b). The data entry block in Fig. 4a consists of three parts for handwritten data collection: isolated digits (F2), alphabet letters (F3), and shapes (F4). In the data entry part of Page #2 of the designed forms, all the participants handwrote (copy) the same specific typewritten text (F7 in Fig. 4b). In data collection process, our 210 volunteers guided their corresponding family members to fill the HSFs. A total of 4,256 Handwriting Sample Forms (HSFs) were completed by all the families/relatives/ and centers, then collected, scanned in a true color format (24 bits per pixels) with 300 DPI resolution, and stored in TIF image format files. The next section describes the HSF data extraction process.

## 3. HSF data extraction

As seen in Fig. 5, the data extraction from the scanned HSF forms was performed in four main steps. The following four subsections briefly explain this process.

### 3.1. Form preprocessing

In order to provide appropriate conditions for the extraction of handwritten fields from the scanned HFSs, the preprocessing step consists of three stages, as illustrated at the top of Fig. 5. Binarization is the first preprocessing step, in which the Otsu method [41] is applied to true color HSFs for deriving a binary version of the forms. In the second step, median filtering (size 3*3) [42] removes the salt and pepper noises on the binary forms. The third preprocessing step performs the estimation of the skew angle ($\theta$) and the skew correction of HSFs forms, which are simply performed by the horizontal and vertical alignment of each HSF's four black corner boxes.

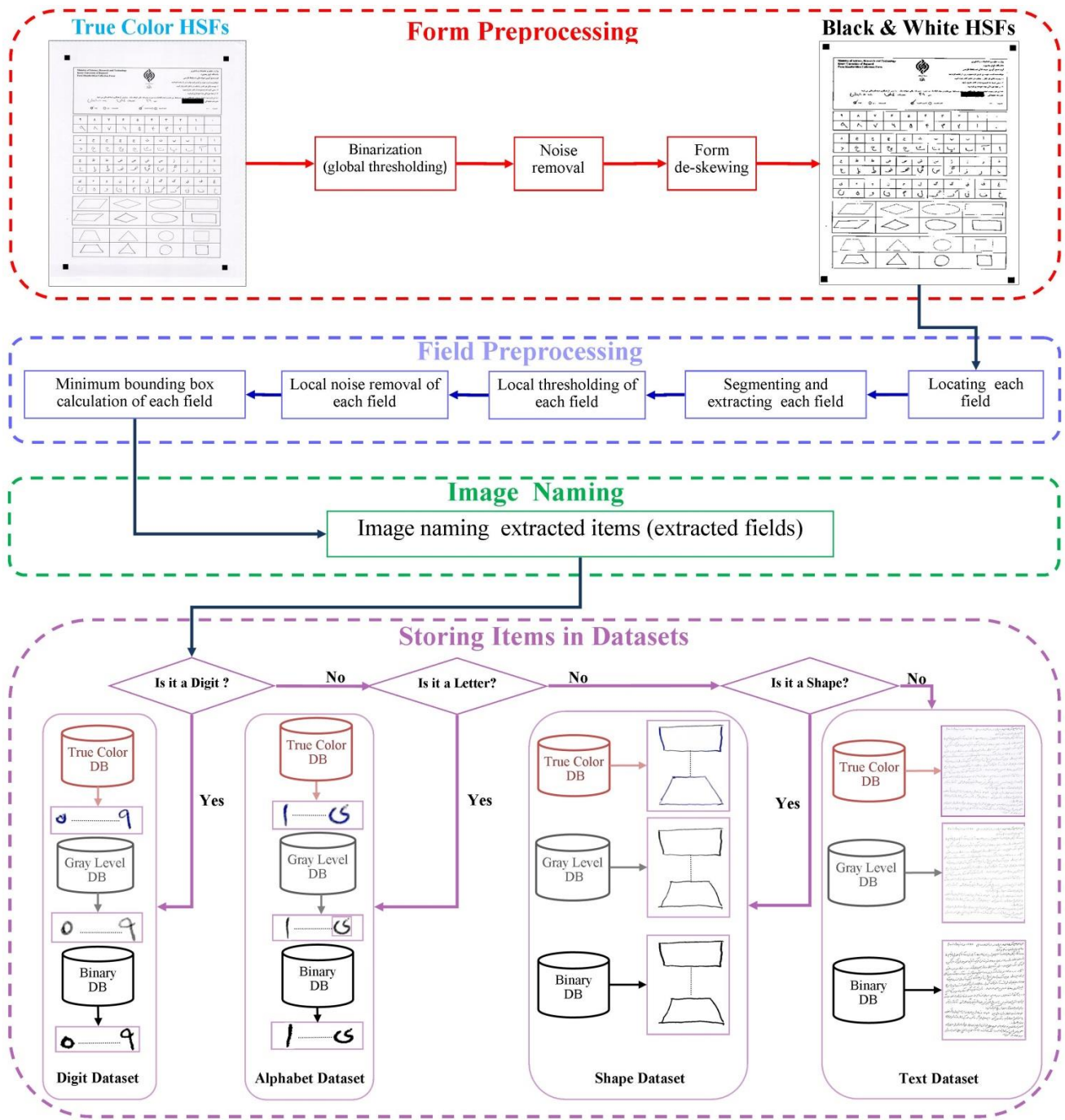

Fig. 5. Overall view of HSF data extraction and data storage in the corresponding dataset.

## 3.2. Field processing

The middle section of Fig. 5 outlines the stages of field processing. The first stage in field processing is to find a location for each handwriting field. The coordinates of the handwritten fields are found after skew correction of the true color forms using θ. In the second stage, each true color field on the forms is segmented and then extracted by utilizing these coordinates to the corresponding true color HSF. As for the third stage, via a gray-level histogram [41], a gray threshold is calculated for each extracted true color. Then, an eye-verification process manually fine-tunes the calculated local threshold in order to improve the quality of the resulting gray field. In the next stage, blobs the size of the less than 10 pixels are identified as noise in each obtained gray image and these are removed. In the last stage, a minimum bounding box is calculated for each segmented field.

## 3.3. Image naming

A unique name is created for each extracted and segmented field. Fig. 6 and Fig. 7 provide the details of name creation for a segmented field in the HSFs, e.g. for field# B43 in P#1. As seen in these figures, each unique name consists of several sections: 1) family unique coding (each family has a unique four-digit code), 2) multiple collection times (indicated by T1, T2 or T3, as shown in Table 2), 3) the gender of each family's center, 4) the family relationship code of the writer with respect to the center, 5) the page number (in this case, P#1), 6) the image field# (in this case, B43). Since centers are the most available persons in this study and they have a unique role in the collection of handwritings of their families, each center filled out his/her two HSFs at three different times, once a month for 3 consecutive months, as indicated by labels of T1, T2, and T3 (Table 2). The rest of the family members, however, completed the HSF forms only once (T1).

Table 2. Multiple collection times for the same writers once a month for 3 consecutive months.

| T1 | Samples of the same writers, written the first month of data collection |
|---|---|
| T2 | Samples of the same writers written the second month of data collection |
| T3 | Samples of the same writers written the third month of data collection |

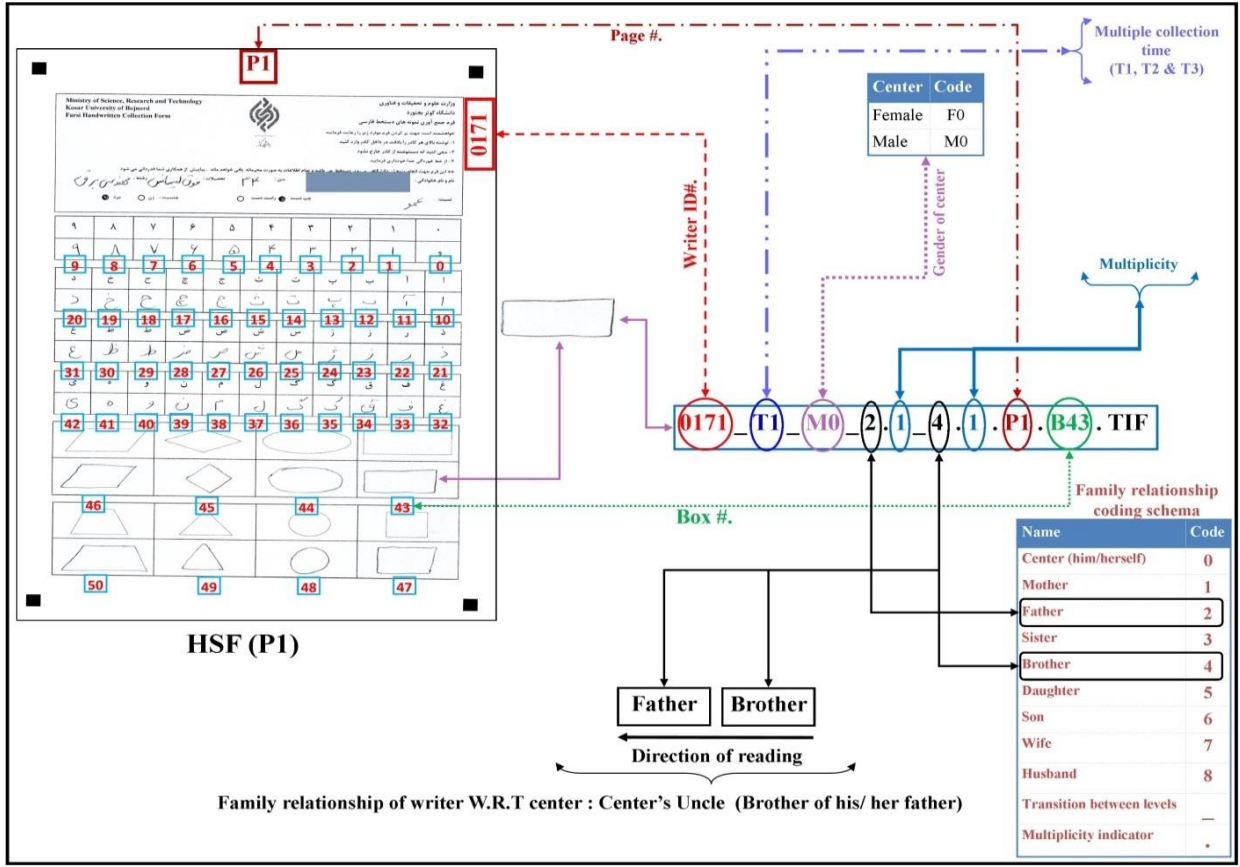

Fig.6. An example of image name creation for the 43rd image field on Page #1.

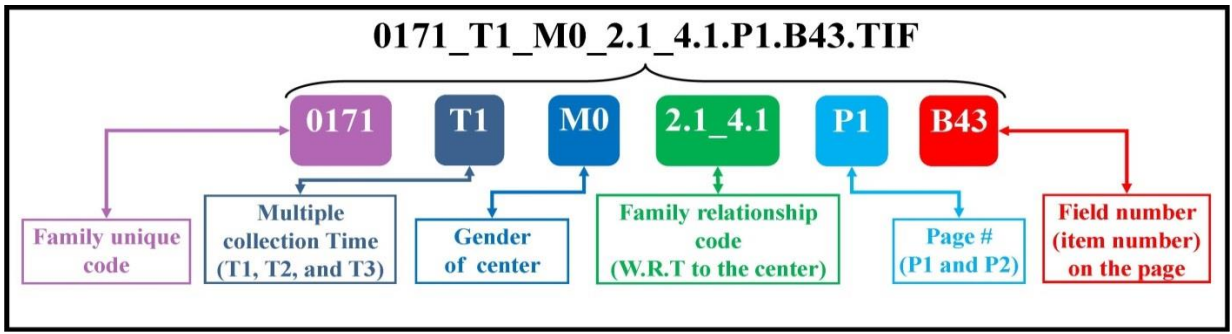

Fig.7. Anatomy of naming an extracted image field.

### 3.4. Storing items in datasets

As seen at the bottom of Fig. 5, the organization of the extracted image fields into proper datasets is the fourth step of the data extraction process. First, for every family, a folder named after the center's name is considered. Then, for each family, three folders: (binary, gray level, and true colors) are created and each extracted image field with a unique name is saved in the corresponding proper folder. The next section presents an overview of the current study's database.

## 4. Database overview

This section presents an overview of the database created. In total, the immediate family members and relatives of 210 families provided the samples of handwritings in our database. In addition, the current study carefully encoded their family relationships and assigned unique names to all their segmented handwritten fields. Fig. 8a illustrates the gender distribution within families. The distribution of handedness for all individuals is shown in Fig 8b. In Fig. 8c, the minimum, maximum, and average age among families are given. Fig.8d provides the distribution of different educational levels for all individuals, from primary to postgraduate school. Fig.8e presents the current work's database of the overall distribution of family relationships of all individuals among all families.

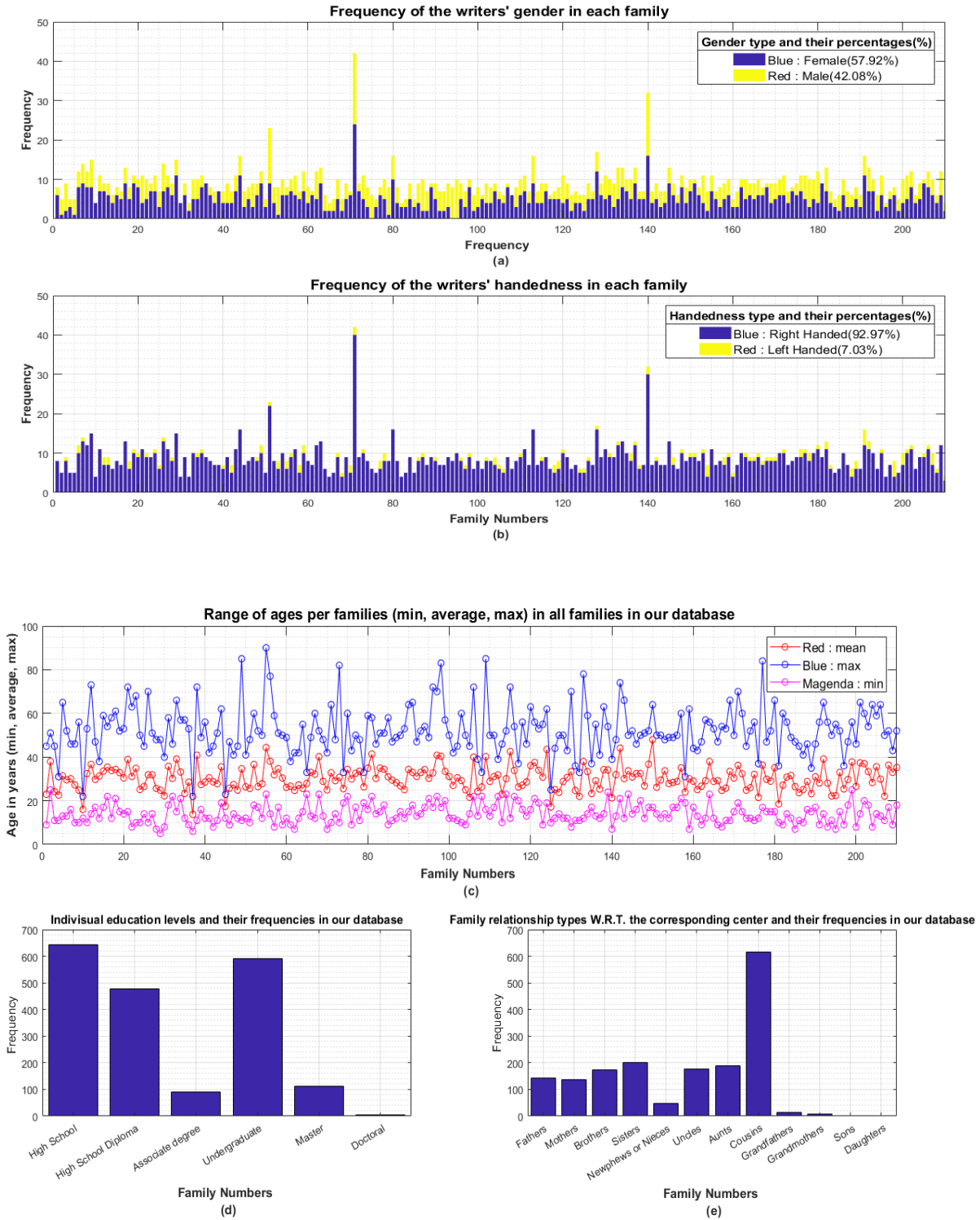

Fig. 8. Some overall statistics: a: distribution of gender among families, b: distribution of handedness among families, c: min, max, and mean age range among families, d: distribution of educational levels among all individuals in all families and, e: distribution of family relationships of all individuals among all families.
16

## 4.1. Database structure

The overall view of this comprehensive database with all of its different folders and image subsets is shown in Fig. 9. As seen, the present work's comprehensive database is comprised of four main datasets (or folders) of written items in Persian handwritten script. These include: digits, alphabetical letters, shapes, and unconstrained free texts. Additionally, the database makes available all 4,256 (=2218*2) HSFs used to collect these. As seen in Fig. 9, all the images of this database have been prepared in three formats (true color, gray level, and binary (black and white)) in order to provide more options to researchers for experimentation. This database also contains Ground-Truth data and metadata, which provide a detailed description of all image contents and writer information in all folders, such as family code, age, handedness, gender, and the writer's relation with respect to the center. Section 4.2 will provide more details about the current work's Ground-Truth information. More elaborate explanations of all the items and the different folders of this database are provided in the following subsections.

### 4.1.1. Digits dataset

Similar to Latin, the digits and numeral strings in Persian are written from left to right. In the present work's database, each digit (0–9) is written one time by each participant (1*10 = 10 digits per writer overall), except for the center. Table 3 presents some samples and their distribution in the database.

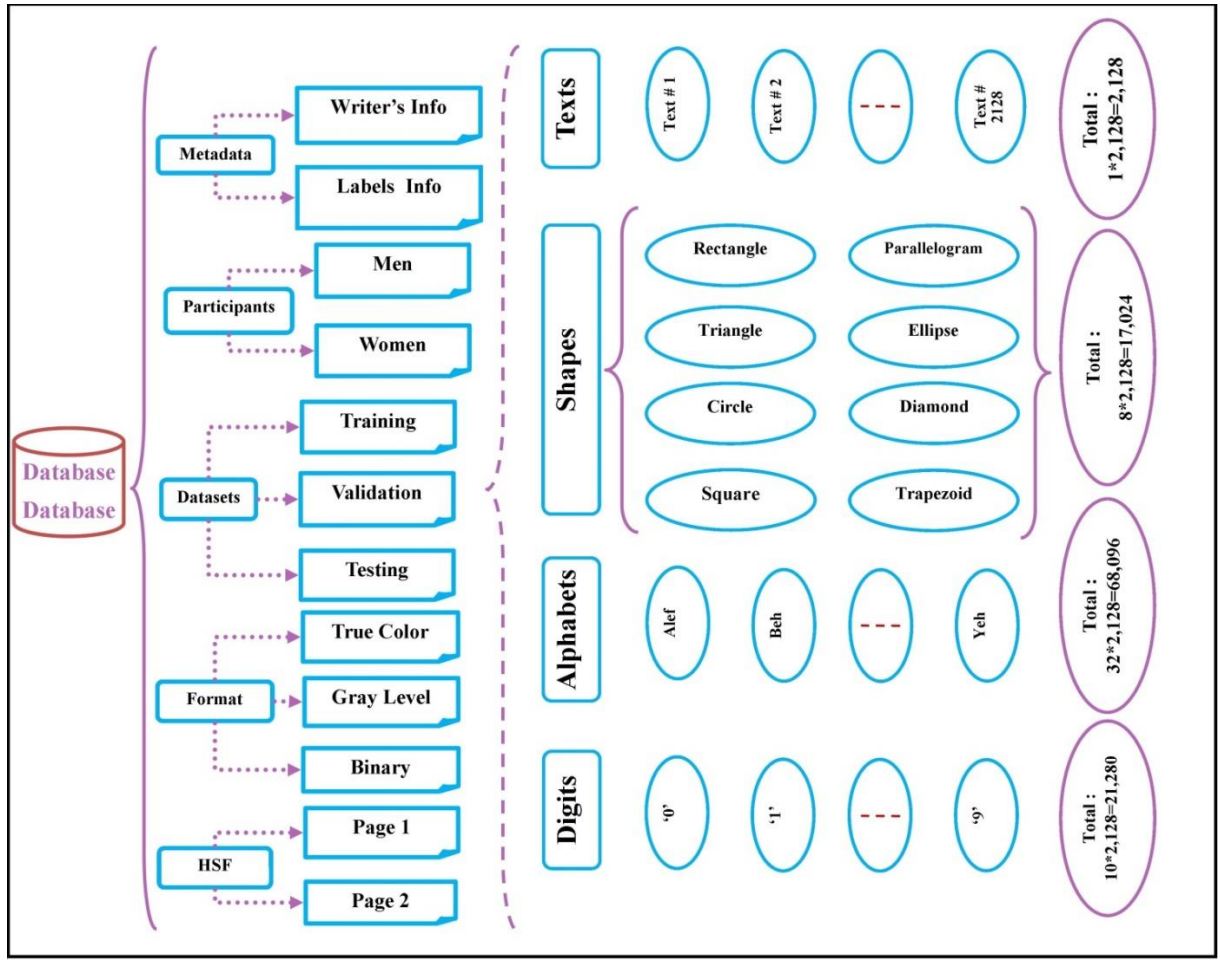

Fig .9. Overall view of the database structure and its four datasets.

Table 3. Farsi isolated digits and their statistics.

| Digit | 0 | 1 | 2 | 3 | 4 | 5 | 6 | 7 | 8 | 9 | Total # of Digits |
|---|---|---|---|---|---|---|---|---|---|---|---|
| Some Samples | ○ | ١ | ۲ | ۳ | ۴ | ۵ | ۶ | ۷ | ۸ | ۹ | |
| | ● | ١ | ۲ | ۳ | ۴ | ۵ | ۶ | ۷ | ۸ | ۹ | |
| | ○ | ١ | ۲ | ۳ | ۴ | ۵ | ۶ | ۷ | ۸ | ۹ | |
| # of Each Digit | 2,128 | 2,128 | 2,128 | 2,128 | 2,128 | 2,128 | 2,128 | 2,128 | 2,128 | 2,128 | 21,280 |

### 4.1.2. Alphabet letters dataset

Persian script consists of an alphabet of 32 letters which is an extension of the Arabic alphabet letters plus four Persian specific letters [43]. In the present study's database, there are 210 families with 2,128 writers and a total of 68,096 (= 32*2,128) letters gathered. Table 4 presents some handwritten letters.

Table 4: Some examples of alphabet letters and their statistics

| General Name | Isolated Form | Some Samples | | | | # of Each Letter |
|---|---|---|---|---|---|---|
| Che | آ | | | | | 2128 |
| Yeh | ش | | | | | 2128 |
| Shin | ق | | | | | 2128 |
| Ghaf | ه | | | | | 2128 |
| He | ی | | | | | 2128 |
| Alef | ج | | | | | 2128 |

### 4.1.3. Shapes

One of the unique properties of the proposed database is its inclusion of geometric shapes which have not yet been featured in Persian handwriting databases. Each writer drew eight shapes, namely the rectangle, ellipse, diamond, parallelogram, circle, square, triangle, and trapezoid, all of which can be utilized in applications such as shape detection and recognition. There are a total of 17,024 (=8*2128) shapes in the current work's database. Table 5 presents some sample shapes and the distribution of each sample in the database.

Table 5: Shape images and overall statistics

| Shape | Some samples | | | # of each shape |
|---|---|---|---|---|
| Rectangle | 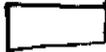 | 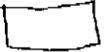 | 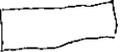 | 2,128 |
| Ellipse | 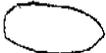 | 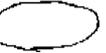 | 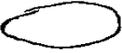 | 2,128 |
| Diamond | 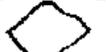 | 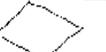 | 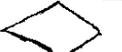 | 2,128 |
| Parallelogram | 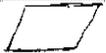 | 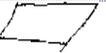 | 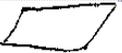 | 2,128 |
| Circle | 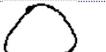 | 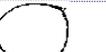 | 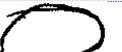 | 2,128 |
| Square | 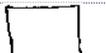 | 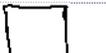 | 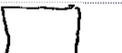 | 2,128 |
| Triangle | 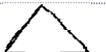 | 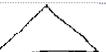 | 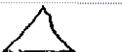 | 2,128 |
| Trapezoid | 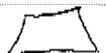 | 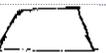 | 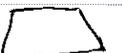 | 2,128 |
| Total # of Shapes | | | | 17,024 |

### 4.1.4. Text database

Free handwritten Persian text by family members is another contribution by the current work's database. A paragraph of handwritten text was designed and collected for the purpose of presenting the variations of digits, numeral strings, letters, symbols, words, and sentences within the context of a paragraph. Each participant freely handwrote (copy) a sample of typewritten text which included 271 words, 5 numeral strings, and some punctuation symbols embedded in the text, all of which demonstrated the variation of these items in the context of a paragraph, lines, and other surrounding words (see Fig. 10). The text presented challenges in the area of handwriting recognition, such as text line segmentation/extraction, baseline detection/correction, segmentation and extraction of numeral strings on text lines from surrounding words, word segmentation, word spotting, word recognition, slant correction, line skew correction, and writer identification and verification. Fig. 10 provides a sample of handwritten text with its corresponding Ground-Truth content. The writer's name has been blocked for confidentiality. In the current study's database, there is a total of 2,128 (= 1*2128 writers) free texts.

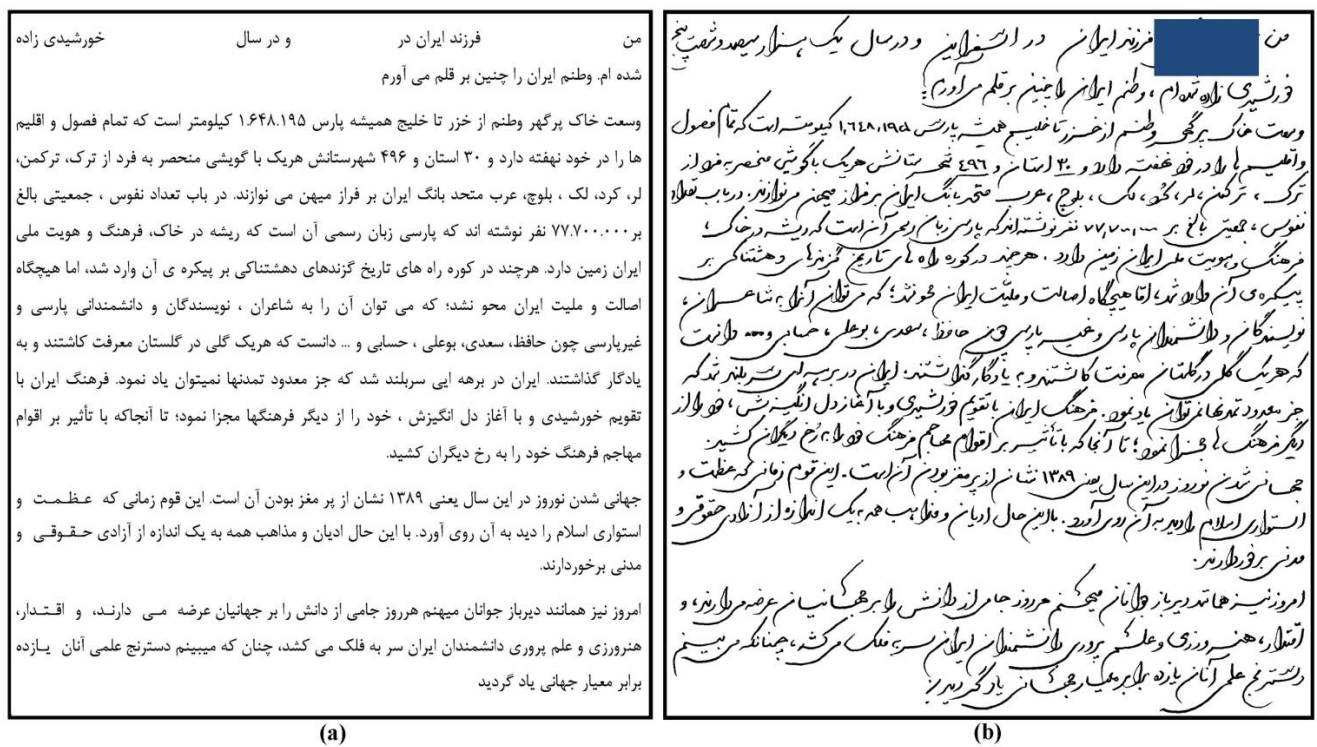

Fig. 10. A binary text sample: (a) Persian typewritten text, (b) Handwritten sample of (a). All the participants freely copied text in (a) by their handwritings.

## 4.2. Metadata and Ground-Truth

Metadata and Ground-Truth files are essential parts of the current study's database. Such a huge database, with its different collections and thousands of items and samples, should contain very rich metadata to facilitate the usage and interpretation of image samples. Unlike other databases, the proposed database also defines every writer based on his/her family and so defines the family relationship Ground-Truth (GT) for each writer. For each family, there are two Microsoft Excel® files that hold all the ground truth (GT) information about every writer and every segmented field written by that writer. As seen in the Table 6.a , each writer in the writer GT file has one row that provides his/her family code, first and last name, gender, relation with respect to the center, educational level, age, and handedness. Also, in the Table 6.b, each image field in the image GT file has one row that provides image file name, multiple collection time, writer family relationship, page#, Field#, image type, content of the segmented image. The GT information in Table 6.b already was illustrated in Fig 7. In the next section, results of some experiments on the proposed database are presented.

Table 6. Two examples of the current study's metadata (Grand-Truths). (a): Writer Ground Truths and (b): Image (Label) Ground Truth.

(a)

| Writer Ground Truth | |
|---|---|
| Name, Family | ▮ |
| Family Unique Code | 0171 |
| Family Relation to the Center | Uncle |
| Gender | Female (F) |
| Age | 34 |
| Handedness | Left-handed(L) |
| Educational Level | Master |

(b)

| Image Ground Truth | |
|---|---|
| Image File Name | 0171_T1_M0_2.1_4.1.P1.B43.TIF |
| Multiple Collection Time | T1 |
| Writer family relationship | 2.1_4.1 |
| Page # | Page #1 (P1) |
| Field # | B43 |
| Image Type | Shape |
| Content of the Image | Rectangle |

## 5. Experimental results

The present work's main goal was to provide a comprehensive database to facilitate investigation of the effects of inheritance and family relationships on handwritings. To the best of our knowledge, no such comprehensive handwritten database has been created for such investigations. In order to demonstrate the usefulness of our newly created database for such researches, in this section, some initial experimental results on the similarity analysis of handwritten texts among family members are presented. It should be emphasized that providing state of the art of handwritten feature extraction and classifications methods, also providing novel techniques for handwriting similarity analysis are not the objectives of the experiments in this section, therefore we have used some classical and well-known features and similarity measures that have been used extensively in the area of handwriting recognition. The main steps of these experiments are preparing data, feature extraction/quantization, and similarity/distance calculations (see Fig. 11). The following subsections discuss the details of these steps.

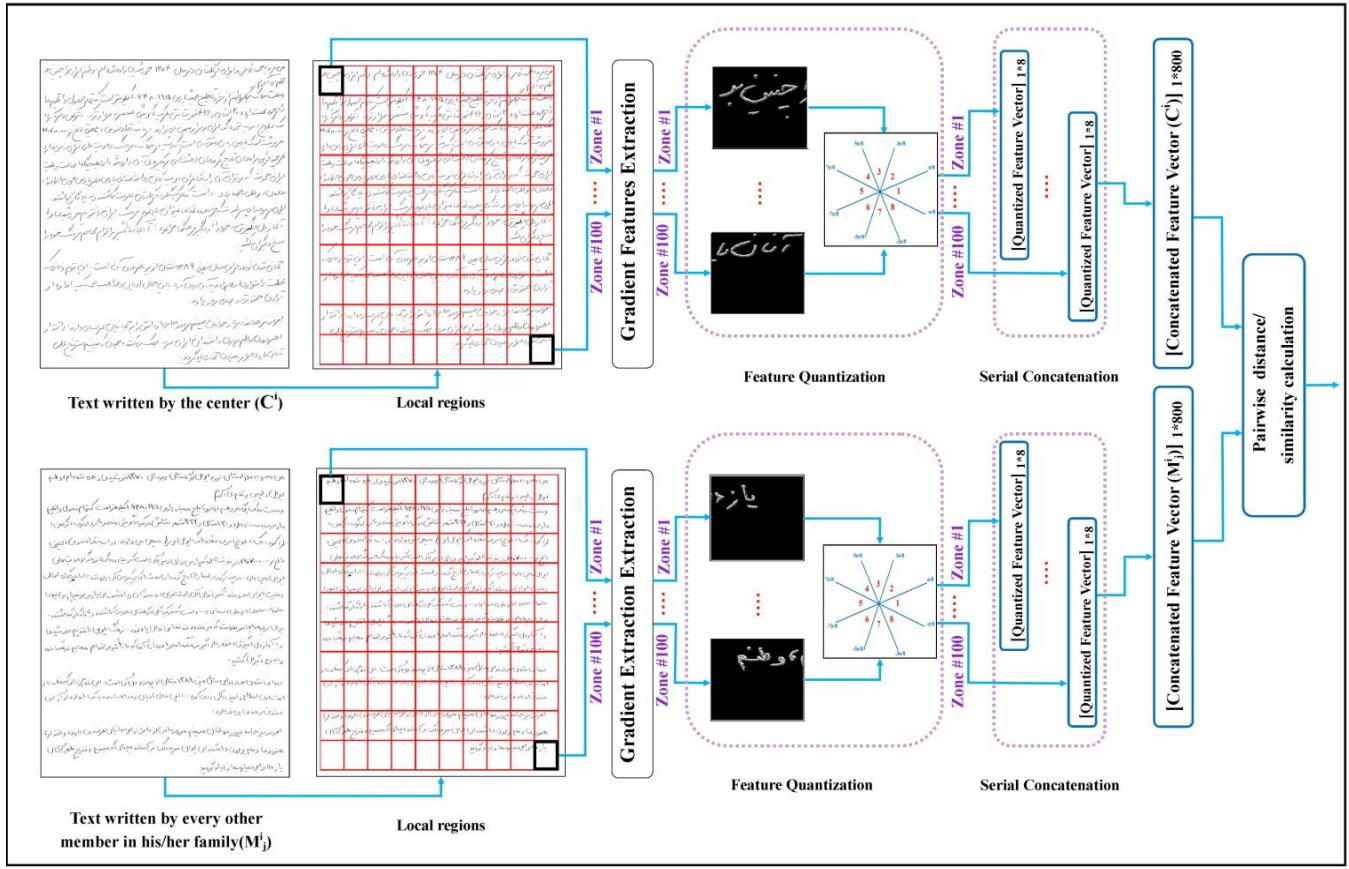

Fig.11. The main steps of the experiments for each family. Here, gray level images of all handwritten paragraphs written by members of the same family are utilized. $C^i$ and $M^i_j$ show center and the $j^{th}$ member of the $i^{th}$ family, respectively. Text images are divided into 100 equal zones and for each zone a quantized feature vector (1*8) is extracted. Then they are concatenated (1*800) and compared.

## 5.1. Data preparation

For these experiments, we have used gray level images of text paragraphs written by all members of all families (Dataset in Section 4.1.4). As shown in Fig. 10, all the participants freely have written the same text in their handwritings. In each family, first images of the text written by its center and its other members are normalized with respect to the largest text size image in the same family, then each normalized text image is divided into some equal regions (zones). After conducting several experiments for determining the size of each zone, text images were divided into 100 zones (=10 vertical *10 horizontal, as seen in Fig. 11).

## 5.2. Feature extraction and quantization

In the literature of document analysis, Directional Gradient Feature (DGF) has shown significant performance among other feature extraction methods [40,44]. The DGF has also demonstrated efficacy in other image recognition problems, such as face and facial recognition[45]. In order to measure the similarity of handwritten paragraphs written by the center ($C^i$) and $j^{th}$ members ($M^i_j$) of the same family, the DFG operator is chosen as the feature extraction method in our experiments. Here the DGF codes the handwriting style of a local region of handwritten text by quantizing the gradient directional angles, as depicted in Fig.12 [40].

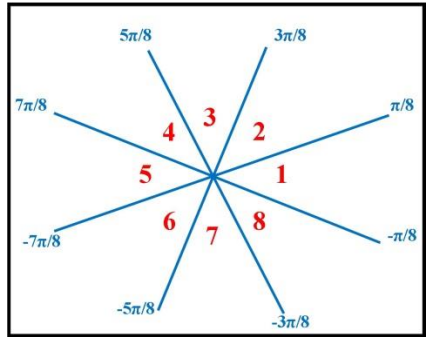

Fig.12. The main eight directions (four orientations) and their quantization intervals of size pi/4 for the DGF.

As seen in Fig.11, in each region (zone), the direction (ө) at each pixel (u, v) is calculated based on Eq.1.

$$\theta(u,v) = tan^{-1}\left(\frac{g_x(u,v)}{g_y(u,v)}\right) \quad (1)$$

where the gradient's horizontal and vertical components, i.e., g$_x$ and g$_y$, are calculated using the Sobel operator, as illustrated in Eqs.2 and 3, respectively[40].

$$g_x(u,v) = I(u-1, v-1) + 2 \times I(u-1, v) + I(u-1, v+1) - I(u+1, v-1)$$
$$-2 \times I(u+1, v) - I(u+1, v+1) \quad (2)$$

$$g_y(u,v) = I(u-1, v-1) + 2 \times I(u, v+1) + I(u+1, v+1) - I(u-1, v-1)$$
$$-2 \times I(u, v-1) - I(u+1, v-1) \quad (3)$$

In Eq.1, ϴ (u, v) shows the direction of gradient vector (g$_x$, g$_y$) in the range of [-π, π]. In our experiment, the gradient direction at each pixel (u, v) is quantized into eight equal intervals of size π/4 each (Fig. 12). Then for each region (zone), the quantized angles are counted, and they are represented by a vector 1*8, afterwards, all these vectors for all 100 regions are concatenated as a 1*800 vector (see Fig. 11). Finally, a simple linear normalization is applied in order to normalize feature values of these vectors between [0, 1] (Eq.4), where $x_f$ is the $f^{th}$ feature value of these feature vectors.

$$F(x_f) = \frac{x_f - \min(x_f)}{\max(x_f) - \min(x_f)} \quad , (f = 1..800) \quad (4)$$

## 5.3. Pairwise distance (similarity) calculation

For comparison of the feature vectors of text images we considered using three classical distance/dissimilarity measures including Euclidian, Cosine, and City Block distances. Among these three distance measures, we chose Euclidian distance (Eq.5) as its distance/similarity values were more consistent with human eyes' verification.

$$Euclidian(F(C^i), F(M_j^i)) = \sqrt{\sum_{l=1}^{800}(F(C^i), F(M_j^i))^2} \quad (5)$$

Using this Euclidian distance (Eq.5), DGF of the handwritten text (paragraphs) written by the center ($C^i$) were compared with feature vector of all other family members of the same family ($M^i_j$). Then we found the most similar (min distance) and the most dissimilar (max distance) text to the center within the same family, and we conducted this experiment for all of our 210 families. Fig.13, Fig.14, Fig.15 and Fig.16 show results for some of these families.

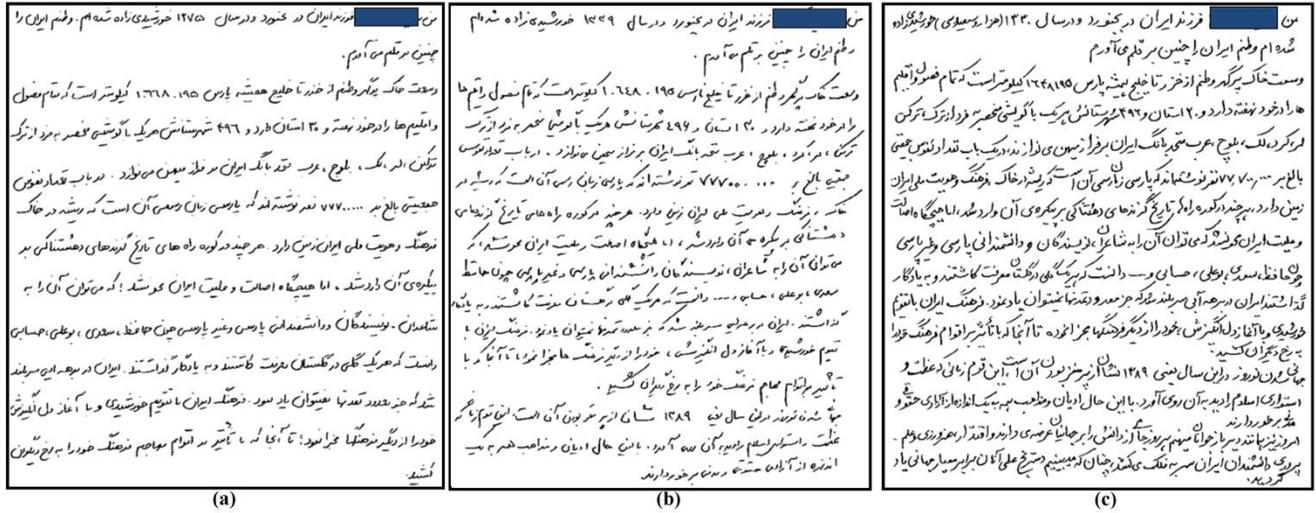

Fig. 13. Three samples of handwritten text from Family#0032. (a) Center: Age = 21, Gender = F, Handedness = R, Education = Undergraduate (b) the most similar: her aunt: age = 55, Gender = F, Handedness = R, Education = Undergraduate, (c) the most dissimilar: her father: Age = 66, Gender = M, Handedness = R, Education = Undergraduate.

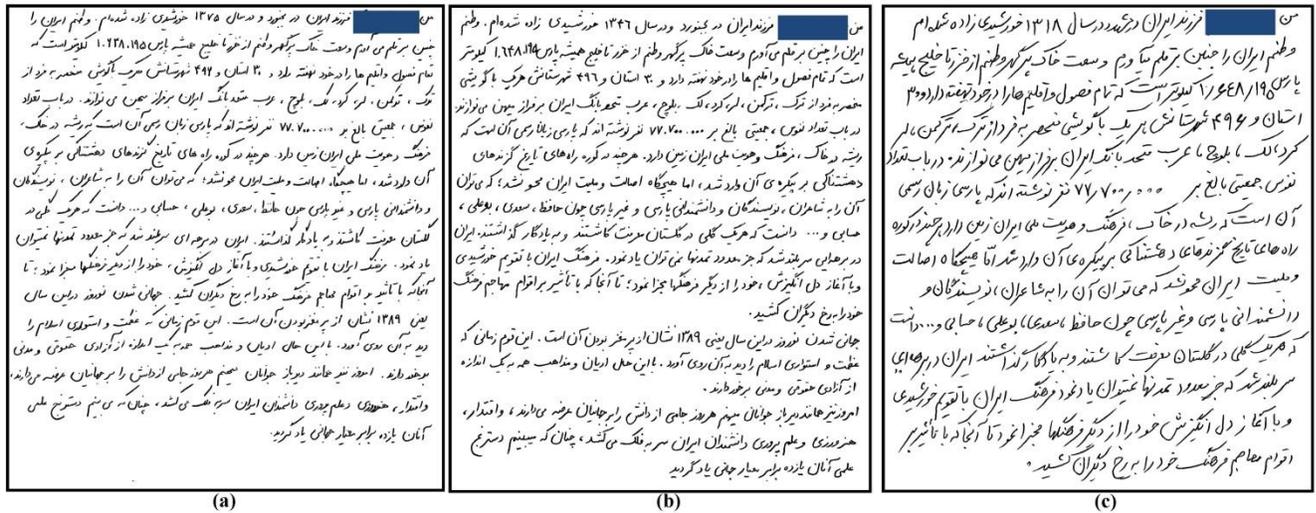

Fig. 14. Three samples of handwritten text from Family#0055. (a) Center: Age = 21, Gender = F, Handedness = L, Education = Undergraduate (b) the most similar: her mother: age = 50, Gender = F, Handedness = R, Education = Undergraduate, (c) the most dissimilar: paternal grandfather: Age = 77, Gender = M, Handedness = R, Education = under the diploma.

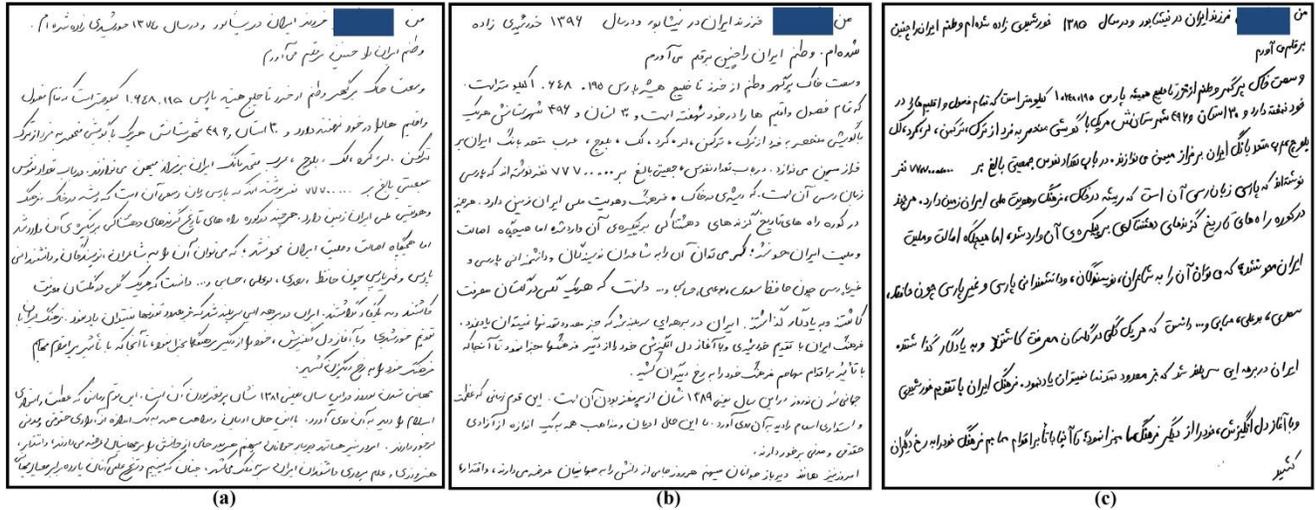

Fig. 15. Three samples of handwritten text from Family#0116. (a) Center: Age = 22, Gender = F, Handedness = L, Education = Undergraduate (b) the most similar: her cousin: age = 19, Gender = F, Handedness = L, Education = Diploma, (c) the most dissimilar: her cousin: Age = 11, Gender = M, Handedness = R, Education = under the diploma.

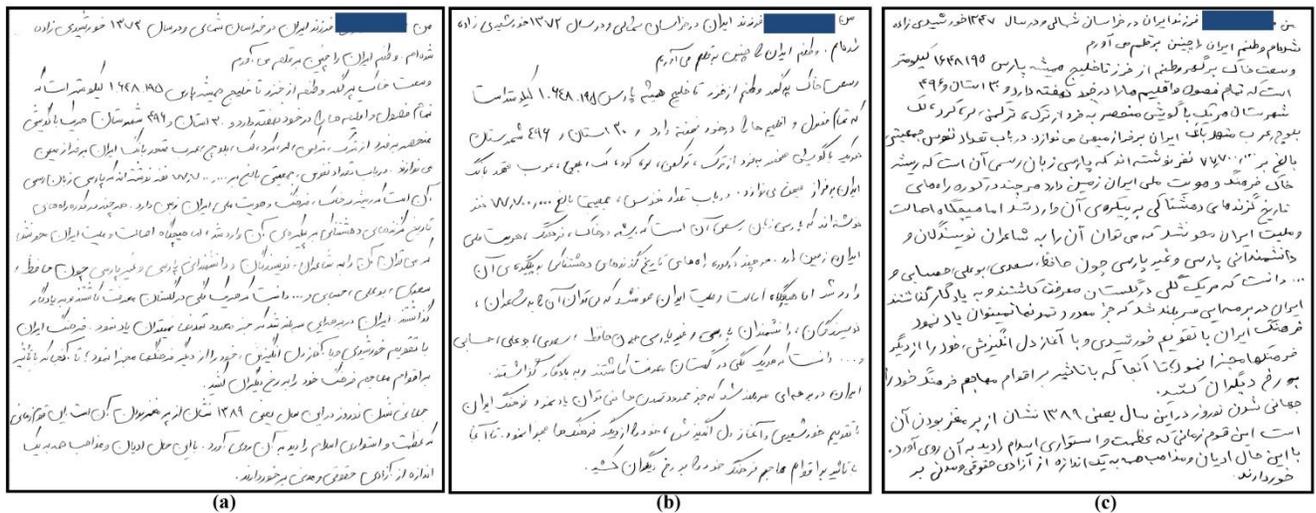

Fig. 16. Three samples of handwritten text from Family#0185. (a) Center: Age = 23, Gender = F, Handedness = R, Education = Undergraduate (b) the most similar: her sister: age = 24, Gender = F, Handedness = R, Education = Undergraduate, (c) the most dissimilar: her father: Age = 50, Gender = M, Handedness = R, Education = under the diploma.

Using more sophisticated feature extraction methods or enriching them by combining other features such as: slant, skew, pen pressure, entropy [23] are out of goal of the current paper, however even using this simple method, within each family we could easily find family members that can copy the handwriting of their respected centers with a very high similarity (or vice versa, as seen in Fig.13-16). Which these results should be interesting for researches in applications such as document examination or writer identification. These experiments show that since in our database all ground truths including all family relationships of the writers and their written components have been captured and associated, these types of investigations and many other experiments are easily possible. According to these results, our database facilitates investigation and association of family relationships with handwritings features, which hopefully it will pave the way for further investigations by other researchers in many important applications in handwritten document recognition areas such as: document examination, writer identification, etc.

## 6. Discussion and comparison

Table 7 shows an overall comparison on distinct aspects of our proposed database with similar state of the art handwritten databases. As seen in this table, our developed database has several unique properties. In our database for the first time family relationships among the writers have fully been considered and recorded in the Ground-Truth information. In total, handwritings of 210 families with 2128 family members were collected. For relationship coding, a novel protocol was introduced. With this new protocol, family relationship is easily encoded by combining some digits and special characters which used in the naming of all the images of handwritten collected items in this database. As seen in Table 7, in comparison with other state of the art databases for different scripts, this database in terms of number of writers, their geographic variability, diversity of participants in terms of age, education and gender, number of families, also in terms of handwritten collected items such as: digits, letters, geometric shapes, paragraphs, modality of images (RGB color images, gray level images, black and white images) is unique. In data collection process to capture variation of handwriting with time, writers (centers of families) were asked to fill out HSFs in multiple times (in three consecutive months, specified by: T1, T2 and T3).

This new database as a comprehensive databases, can be used for solving wide range of challenges in pattern recognition, and machine learning, and handwritten recognition community such as: handwritten character/ word recognition, word spotting, word/ line segmentation, text binarization, skew and slant detections and corrections, shape recognition, gender/age/handedness detection, feature extraction, clustering, document question examination, writer identification/ verification, numerical strings detection and extraction from text for creating important applications like mail sorting, general handwriting recognition software, etc. Also for the first time, it creates new opportunities for researchers to investigate the effects of inheritance and family relationships on handwriting. This database is provided freely to all researchers in the research community upon request[1].

## 7. Conclusion and future works

The effects of inheritance and family relationships on handwritings have not been investigated thoroughly and still some questions are unanswered such as: does heredity affects handwriting? Do family relationships affect handwriting? This paper paves the way and introduces a compressive publicly available database as the first step towards such investigation. To the best of our knowledge, no such database currently is available. In total, a rich collection of samples of handwritings of 210 families with 2128 family members (writers) were collected using especially designed handwritten sample forms (HSFs), and family relationships among all writers within families were fully codded and recorded based on a novel coding protocol.

---

[1] A sample version of this database can be downloaded through the following link:
https://users.encs.concordia.ca/~j_sadri/HeritabilityHandwritingDatabase.html

Table 7.Comparison between our novel database with some common available handwritten databases

| Database name | Language/ script | No. of families | Writers' family relationships | No. of writers | Isolated Digits | Alphabet letters | Shapes | Texts/ Paragraphs | Multiple collection time |
|---|---|---|---|---|---|---|---|---|---|
| **CEDAR** [28] | English | ----- | No | 1500 | 21,179 | 27,837 | ----- | 1500 | No |
| **IAM** [30] | English | ----- | No | 657 | ----- | ----- | ----- | 1539 | No |
| **CVL**[46] | English/ German | ----- | No | 311 | ----- | ----- | ----- | N/A | No |
| **QUWI**[47] | Arabic/ English | ----- | No | 1017 | ----- | ----- | ----- | 5085 | No |
| **RIMES**[48] | French | ----- | No | 1300 | ----- | ----- | ----- | 12723 | No |
| **GRUHD** [49] | Greek | ----- | No | 1000 | 123,256 | ----- | ----- | ----- | No |
| **Hit-Mw** [31] | Chinese | ----- | No | 780 | ----- | 186,444 | ----- | ----- | No |
| **CENPARMI** [50] | Arabic | ----- | No | 328 | 46,800 | 21,426 | ----- | ----- | No |
| **AHTID/MW**[51] | Arabic | ----- | No | 53 | ----- | ----- | ----- | ----- | No |
| **KHATT** [35] | Arabic | ----- | No | 1000 | ----- | ----- | ----- | 6000 | No |
| **CENPARMI** [52] | Dari | ----- | No | 200 | 28,000 | 7400 | ----- | ----- | No |
| **CENPARMI**[39] | Persian | ----- | No | 175 | 18,000 | 11,900 | ----- | ----- | No |
| **FHT** [37] | Persian | ----- | No | 250(163+92) Men and Women | ----- | ----- | ----- | 1000 | No |
| **IFHCDB** [53] | Persian | ----- | No | N/A | 17,740 | 52,380 | ----- | ----- | No |
| **HODA** [38] | Persian | ----- | No | N/A | 102,357 | ----- | ----- | ----- | No |
| **IAUT/PHCN** [54] | Persian | ----- | No | 380 | ----- | ----- | ----- | ----- | No |
| **IFN/Farsi-database**[55] | Persian | ----- | No | 600 (340+260) Men and Women | ----- | ----- | ----- | ----- | No |
| **CENPARMI**[56] | Persian | ----- | No | 400 | 24,121 | ----- | ----- | ----- | No |
| **Sadri and et al**[40] | Persian | ----- | No | 500 (250+250) Men and Women | 97,124 | 42,500 | ----- | 500 | No |
| **PHBC-database**[57] | Persian | ----- | No | 900 | 20,628 | ----- | ----- | ----- | No |
| **Our database** | **Persian** | **210** | **Yes** | **2,128(894 + 1234) Men and Women** | **21,280** | **68,096** | **17,024** | **2,128** | **Yes** |

This database includes samples of handwritten digits, alphabets, geometric shapes and free texts/paragraphs with complete Ground-Truth and comprises three image modalities: True Color (RGB) images, gray level images as well as binary images. This database can be used for wide range of studies and researches such as: handwritten character/ word recognition, word/ line segmentation, skew and slant detections and corrections, shape recognition, gender/ age/ handedness detection, document individuality, document question examination, writer identification and verification as well as investigation of the effects of inheritance and family relationships on handwriting. We hope this comprehensive database will extend research in the pattern recognition community and will answer questions regarding the effects of inheritance on handwriting. In the future, by using tablets and digital pen we would like to provide an online version of our database which will allow us to extract online features from handwritings in order to extend this investigation and enrich our features using online information.

## Acknowledgement

The authors of this paper would like to sincerely thank all 210 volunteers that played the role of centers for their families and patiently collected HSFs from all of their family members and relatives. Without their help creating such a database would not be possible.